# Impact of Emoji Exclusion on the Performance of Arabic Sarcasm Detection Models

A Preprint

**Ghalyah H. Aleryani**
Department of Computer Science in Jamoum
Umm Al-Qura University
Makkah, Saudi Arabia
*S44280249@st.uqu.edu.sa

**Wael Deabes**
Department of Computational, Engineering,
Mathematical Sciences (CEMS)
Texas A&M University-San Antonio
San Antonio, USA
wdeabes@tamusa.edu

**Khaled Albishre**
Department of Computer Science in Jamoum
Umm Al-Qura University
Makkah, Saudi Arabia
kmbishre@uqu.edu.sa

**Alaa E. Abdel-Hakim**
Department of Computer Science in Jamoum
Umm Al-Qura University
Makkah, Saudi Arabia
adali@uqu.edu.sa

May 2, 2024

## Abstract

The complex challenge of detecting sarcasm in Arabic speech on social media is increased by the language's diversity and the nature of sarcastic expressions. There is a significant gap in the capability of existing models to effectively interpret sarcasm in Arabic, which mandates the necessity for more sophisticated and precise detection methods. In this paper, we investigate the impact of a fundamental preprocessing component on sarcasm speech detection. While emojis play a crucial role in mitigating the absence effect of body language and facial expressions in modern communication, their impact on automated text analysis, particularly in sarcasm detection, remains underexplored. We investigate the impact of emoji exclusion from datasets on the performance of sarcasm detection models in social media content for Arabic as a vocabulary-super rich language. This investigation includes the adaptation and enhancement of AraBERT pre-training models, specifically by excluding emojis, to improve sarcasm detection capabilities. We use AraBERT pre-training to refine the specified models, demonstrating that the removal of emojis can significantly boost the accuracy of sarcasm detection. This approach facilitates a more refined interpretation of language, eliminating the potential confusion introduced by non-textual elements. The evaluated AraBERT models, through the focused strategy of emoji removal, adeptly navigate the complexities of Arabic sarcasm. This study establishes new benchmarks in Arabic natural language processing and presents valuable insights for social media platforms.

## 1 Introduction

The evolution of social media platforms has transformed them into a place of free speech, offering users a space to express their ideas and opinions freely. While this open environment motivates useful discussions, it can also become a problem when individuals employ expressions and statements, that may offend others due to diverse beliefs, backgrounds, genders, or races. Although it is protected under the latest version of the Communications Decency Act

*Corresponding author

(CDA 230) [Dumas, 2022], many social media platforms are actively working to enhance user protection against hateful and offensive content. A significant challenge, however, was because of their dependence on human monitoring and user reports [Baggs, 2021, Sukhavasi and Dondeti, 2023].

In recent years, there has been a significant increase in the prevalence of sarcasm speech on social media, giving rise to serious social problems including social conflicts, racist crimes, and the spread of negative social influence. To mitigate this issue, many social media platforms have implemented word filters based on NLP techniques. Sarcasm speech includes exchanges of sarcastic or offensive remarks between individuals, and it extends to text comments to include multimedia content such as videos and audio [Mihi et al., 2021]. Existing research efforts have primarily focused on proposing models for the automatic detection of sarcasm speech within text comments on platforms like Twitter or YouTube [García-Díaz et al., 2023, Koroteev, 2021]. These models typically involve several stages, including data processing, representation, and classification. However, these stages have several challenges, particularly concerning the Arabic language [Farha and Magdy, 2020, Rahma et al., 2023]:

1. Its rich vocabulary, and dialectical variations.
2. Arabic sarcasm often relies heavily on contextual indications and cultural references.
3. Collecting and annotating a large dataset for Arabic sarcasm detection produces unique difficulties.
4. Time-consuming and requires extensive data learning.
5. The multimodal nature of social media content is integrating information from various sources such as text, images, videos, and emojis.

Generally, sarcasm detection heavily relies on textual data classification. However, textual data lacks crucial expressive features such as facial expressions, body language, and tone variations. To address this limitation, social media communities have introduced emojis as non-traditional vocabularies to compensate for the deficiency in information. Emojis have demonstrated their effectiveness in partially bridging the gap between textual and vocal/visual communication [Subramanian et al., 2019].

Nevertheless, rich languages like Arabic possess their own tools that can better address this gap than emojis, including a vast and diverse lexicon. Table 1 provides a comparison of the number of roots between Arabic and other languages. Additionally, Arabic speakers employ numerous dialects with significant dialectical variations. Moreover, sarcasm in Arabic heavily relies on contextual cues, puns, and euphemisms. This results in an extensive textual information space generated solely using textual vocabularies, decreasing the contribution of the limited emoji dictionary.

These factors raise questions regarding the added value of emojis to classifier performance. While it is intuitive that adding redundant data should not degrade performance, excess data has been shown to harm classifiers' performance in certain instances [Abdel-Hakim and Deabes, 2017]. Hence, we hypothesize that using emojis for sarcasm detection in rich languages like Arabic may either reduce accuracy or offer no improvement. The rationale behind this approach is that focusing purely on the textual elements allows the models to concentrate on linguistic and semantic analysis without the potential confounds introduced by emojis.

The successful development of an Arabic sarcasm detection model, enhanced by transfer learning methods and refined by the strategic removal of emojis from the dataset, will have a significant social impact. This approach will enable social media platforms to more effectively detect and moderate sarcastic speech, thereby mitigating its potentially harmful effects and fostering a more positive and respectful online discourse. By focusing on the textual content and reducing the noise in the data, the model's precision in identifying sarcasm will be improved, making the moderation process more efficient and reliable. Additionally, this research will lay the groundwork for further advancements in the creation of language-specific models for sarcasm detection, equipping diverse language communities with the tools needed to address such speech more effectively.

In this work, we investigate the following research questions:

1. Does including emojis in the data improve pre-trained models' ability to detect sarcasm in the Arabic language?
2. How can the performance of the AraBERT pre-trained models for sarcasm detection in the Arabic language on social media platforms be improved by removing emojis from the data?
3. How accurately can it identify and classify sarcasm in Arabic speech on social media?

In the next section, we discuss related work on detecting textual aggressions on social media. In Section 3, we describe the methodology used in this study. Section 4 presents the experiments and analysis of the results. Finally, Section 5 concludes the study and discusses the future direction of sarcasm speech in the Arabic language.

Table 1: Comparison between Arabic and some other languages in terms of the number of roots.

| Language | Approximate Number of Roots | Notes |
|---|---|---|
| Arabic | 23090 [Othman et al., 2020] | Roots of 3 letters<br>By applying 73 triliteral patterns and 18 affixes produced around 27.6M words. |
| English | 8,400 [Biemiller and Slonim, 2001] | This study highlights the significant growth in root word vocabulary during the primary school years. |
| Russian | 450 [Patrick, 1989] | The study highlights the significance of understanding root words for mastering Russian vocabulary. |

## 2 Related Work

User-created content online, especially on social media platforms, can sometimes contain harmful language and hate speech, which has detrimental effects on the minds of the online community and could potentially result in hate crimes. Recently, there is been a marked interest in smart algorithms designed to identify and flag such offensive language and hate speech automatically. Nonetheless, NLP research concerning the Arabic language is typically limited [El-Melegy et al., 2019], as is the investigation into sarcasm speech. However, in this section, we try to highlight previous efforts focused on detecting Arabic sarcasm speech on social media.

In the field of NLP, large-scale pre-trained models have become the standard approach for a wide range of tasks. Models like Bidirectional Encoder Representations from Transformers (BERT) are trained on massive datasets, allowing them to generalize their knowledge effectively to various downstream tasks [Koroteev, 2021]. The BERT model has been employed for extensive Arabic datasets producing AraBERT [Antoun et al., 2020]. In [Althobaiti, 2022], an automated approach to detect offensive language and detailed hate speech in Arabic tweets is proposed. The BERT model is utilized with two traditional machine learning methods: (i) Support Vector Machine (SVM) and (ii) Logistic Regression (LR). Additionally, the authors explore the integration of sentiment analysis and the descriptions of emojis as supplementary features to the textual content of tweets. Experimental results indicate that the BERT-based model outperforms existing benchmark systems in three key areas: (a) detecting offensive language with an F1-score of 84.3%, (b) identifying hate speech with an 81.8% F1-score, and (c) discerning detailed categories of hate speech (like race, religion, social class, etc.) with a 45.1% F1-score. While sentiment analysis marginally boosts the model's efficiency in detecting offensive language and hate speech, it doesn't enhance the model's capability in classifying specific hate speech types. Moreover, a universal language-agnostic approach is proposed in [Mubarak et al., 2023] to gather a substantial proportion of tweets with offensive and hate content, irrespective of their subjects or styles. The authors gathered a significant collection of offensive tweets by leveraging the non-verbal cues in emojis. Then, they apply the proposed methodology to Arabic tweets and juxtapose it with English ones, highlighting on notable cultural variances. A consistent pattern emerged with these emojis signifying offensive content over various periods on Twitter. They hand-label and make publicly available the most extensive Arabic dataset that encompasses offensive language, detailed hate speech, profanity, and violent content. As a result, the authors found that even advanced transformer models can overlook cultural contexts, backgrounds, or precision inherent in authentic data, such as sarcasm.

As highlighted by [Husain and Uzuner, 2022], identifying offensive language within Arabic content is intricate. Several challenges arise such as: (i) The colloquial language frequently used on social media platforms. These posts often contain abbreviations and slang, making it challenging for classifiers to understand and process them semantically. (ii) The diverse dialects and versions of the Arabic language add another layer of complexity to discerning offensive content. The text may require extensive preprocessing before being fed into a classification model. To combat the issue of colloquialism, the researchers process each tweet. This involves translating emoticons and emojis into their Arabic textual equivalents and breaking down hashtags into individual words separated by spaces. Addressing the issue of dialect variation, the texts are transformed from regional dialects to Modern Standard Arabic (MSA). The study tested various classifiers, including traditional machine learning models SVM, LR, Decision Tree (DT), bagging, AdaBoost, and Random Forest (RF). The results show that, among traditional machine learning models, SVM topped the list with an F1 score of 82%, followed by LR at 81% and DT at 69%. For ensemble models, bagging led with an F1 score of 88%, then RF at 87%, and AdaBoost at 86%.

For this study [Al-Dabet et al., 2023], features were derived from textual descriptions of emojis. Depending on the class size and the specific emoji, these features either enhanced or hindered the model's performance. The authors introduce a transformer-based technique to tackle the problem of detecting offensive language. Their approach utilizes variations of the CAMeLBERT model and is tested using a combination of four benchmark Arabic Twitter datasets, all annotated for hate speech detection, including the dataset from the (OSACT5 2022) workshop shared task. The model demonstrated proficiency in identifying offensive content in Arabic tweets, achieving an accuracy of 87.15% and an F1 score of 83.6%.

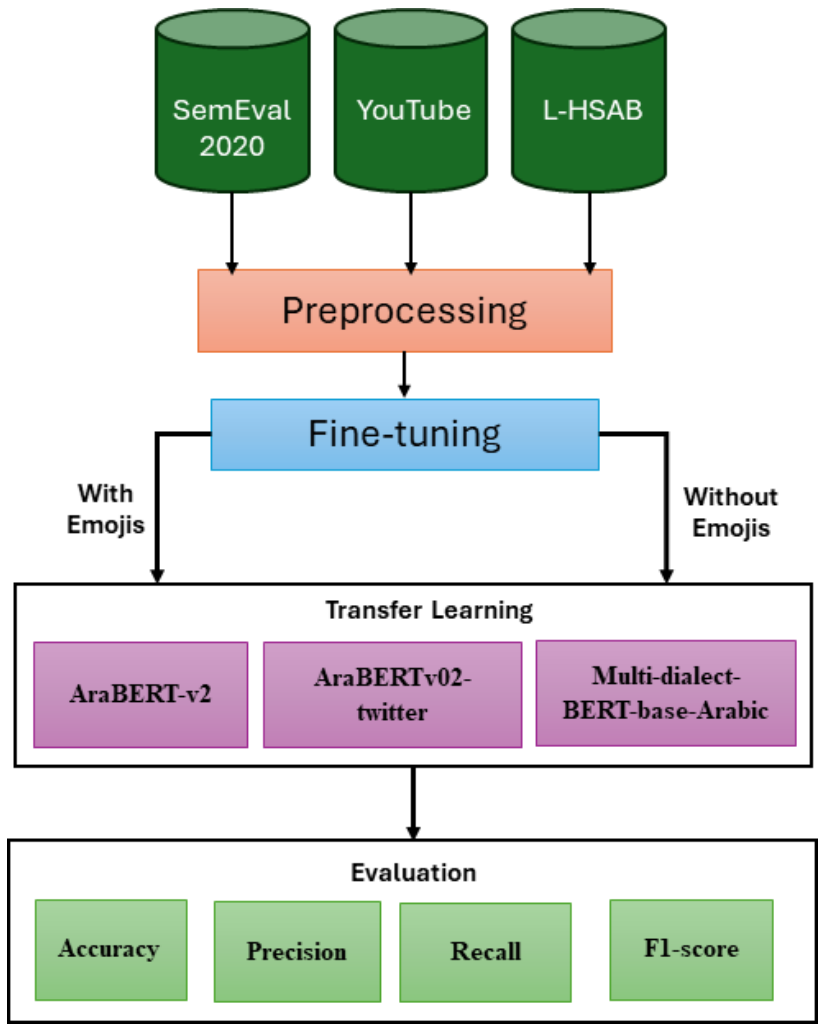

Figure 1: Workflow of the methodology.

In [Ibrohim et al., 2019], the word embedding (word2vec) feature is applied in tandem with part-of-speech and/or emoji to detect such language in Indonesian tweets on Twitter. They also experimented with combining unigrams with part-of-speech and/or emojis. Classification for this study was performed using a Support Vector Machine, Random Forest Decision Tree, and Logistic Regression methods. The highest accuracy attained was 79.85%, with an F-Measure of 87.51%, using the fusion of unigram features, part-of-speech, and emoji.

Furthermore, the work in [Al-Azani and El-Alfy, 2018] investigates the impact of combining emoji-based elements, which are progressively common on social media, with multiple textual factors for the sentiment analysis of casual Arabic tweets. The authors employed four different methods to extract textual features: Bag-of-Words (BoW), Latent Semantic Analysis, and two Word Embedding variants. The study evaluates the impact of merging emojis with these textual elements using the SVM classifier, considering both scenarios: with and without feature selection. Results indicate that models incorporating emojis with word embedding and optimal feature selection produce better outcomes. Further, the implications of amalgamating emoji-based aspects from Arabic tweets with diverse textual elements are explored.

In this study, we hypothesize that the removal of emojis from the training dataset will enhance the ability of AraBERT pre-trained models to discern the subtleties of sarcastic language in Arabic text, as it will encourage the model to focus more on context.

## 3 Proposed Work

We use transfer learning with three pre-trained models: AraBERT-v2, AraBERTv02-twitter, and Multi-dialect-BERT-base-Arabic. The models are evaluated using datasets from three different sources: SemEval 2020, YouTube, and L-HSAB, preprocessing. The evaluation of the models is performed with and without emojis. This suggests that the role of emojis in understanding the text is considered a variable in the model's performance. Finally, the performance of these models is evaluated using standard performance metrics: accuracy, precision, recall, and F1-score.

### 3.1 Classification Model

The core of this study revolves around the selection and application of advanced machine learning models, specifically tailored for processing Arabic text, the workflow of this study is presented in Figure 1.

These models are critical in accurately detecting sarcasm in Arabic social media content. The selected models are:

1. Arabert_v2: This model is a BERT-based framework specifically adapted for Arabic text. AraBERT's architecture allows it to understand the contextual nuances of the Arabic language, making it an ideal choice for tasks like sarcasm detection where context plays a crucial role. Moreover, it comprises 12 transformer layers and 768 hidden units in each layer [Elfaik et al., 2021].

Table 2: Main hyperparameters for fine-tuning the three models.

| Parameter | Value |
|---|---|
| Adam optimizer | 1e-8 |
| Learning Rate | 5e-5 |
| Batch Size | 16 |
| Maximum Sequence Length | 256 |
| Epochs | 15 |

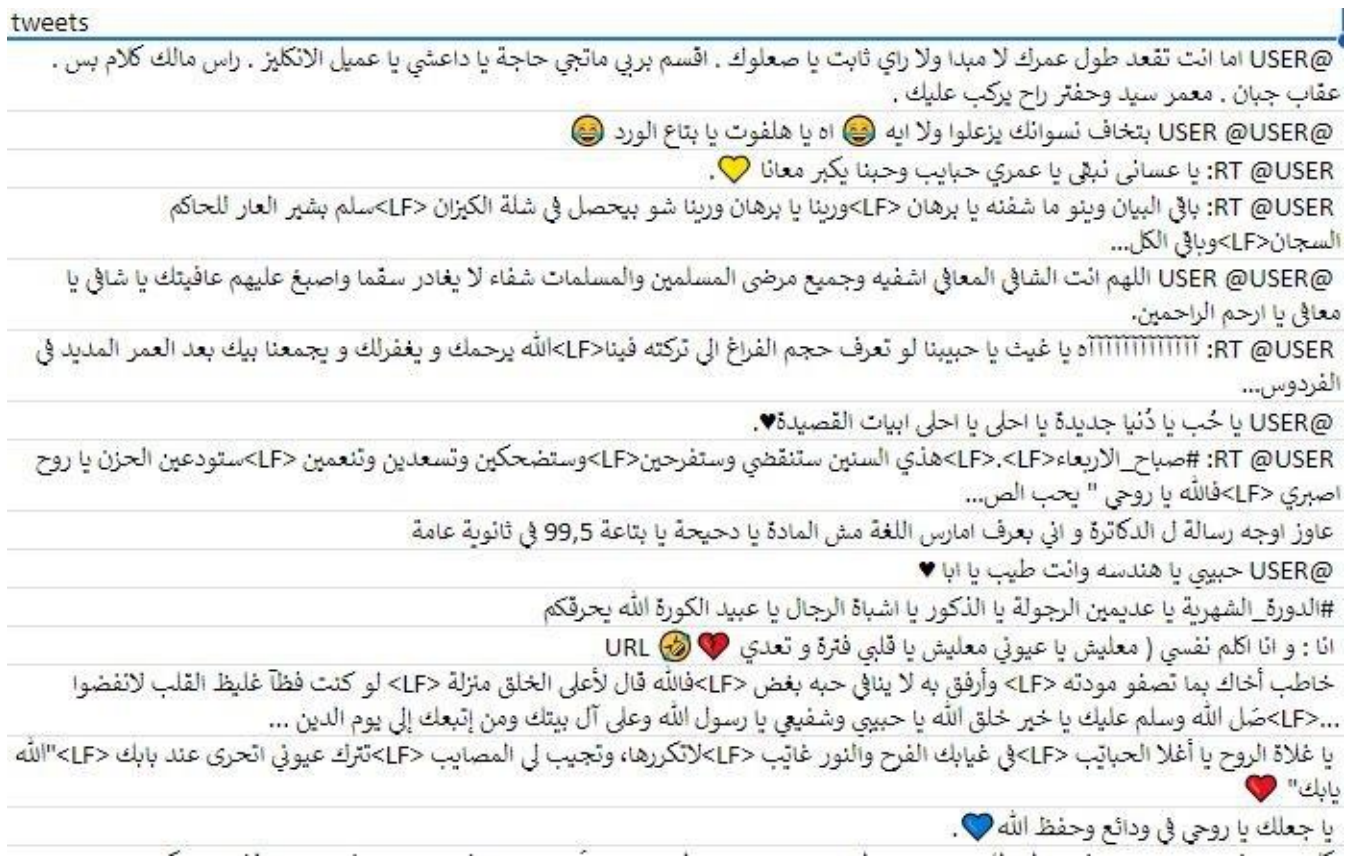

Figure 2: Snapshot of SemEval dataset.

2. AraBERTv02-twitter: Building upon the foundation laid by AraBERT, Arabert_v2 is an advanced version that offers improved capabilities. Its enhanced features include a better understanding of dialectal variations and more refined contextual analysis. This version is particularly effective in handling the intricate aspects of sarcasm in various forms of Arabic language and dialects, maintaining the same architecture layer of Arabert_v2 [Humayun et al., 2023].

3. Multidialect Bert base AraBERT: Recognizing the diversity of the Arabic language, this model is designed to handle multiple dialects. Its architecture is tailored to adapt to the linguistic variations found across different Arabic-speaking regions and the model is trained on the entire Wikipedia for each language, which is crucial for a comprehensive sarcasm detection tool that can operate effectively across diverse social media platforms and content. The architecture is composed of 12 encoder blocks, each equipped with 12 self-attention heads, and is followed by hidden layers that have a size of 768 [Alammary, 2022, Humayun et al., 2023].

Table 2 presents the key hyperparameters for the three models, which have been determined based on empirical.

### 3.2 Datasets

The success of transfer learning models in NLP tasks heavily relies on the quality and relevance of the datasets used for training and testing. In this study, three distinct datasets have been carefully selected to ensure comprehensive coverage of the various facets of Arabic sarcasm speech and related nuances in social media contexts. These datasets are:

1. SemEval 2020 Task 12 Dataset: This dataset is specifically designed for sarcasm detection, making it highly relevant for this research. Figure 2 shows a sample of the SemEval dataset. It comprises a collection of social media posts annotated for sarcasm, providing a foundational basis for training and testing the models. Including this dataset is crucial as it offers direct insights into the textual characteristics and markers indicative of sarcasm in social media content [Zampieri et al., 2020].

2. YouTube Dataset for Anti-Social Behavior Detection: Recognizing the importance of context in sarcasm detection, this dataset from YouTube, comprising comments and posts in Arabic, is employed to understand and identify patterns of anti-social behaviour, as shown in Figure 3 [Alakrot et al., 2018]. This dataset

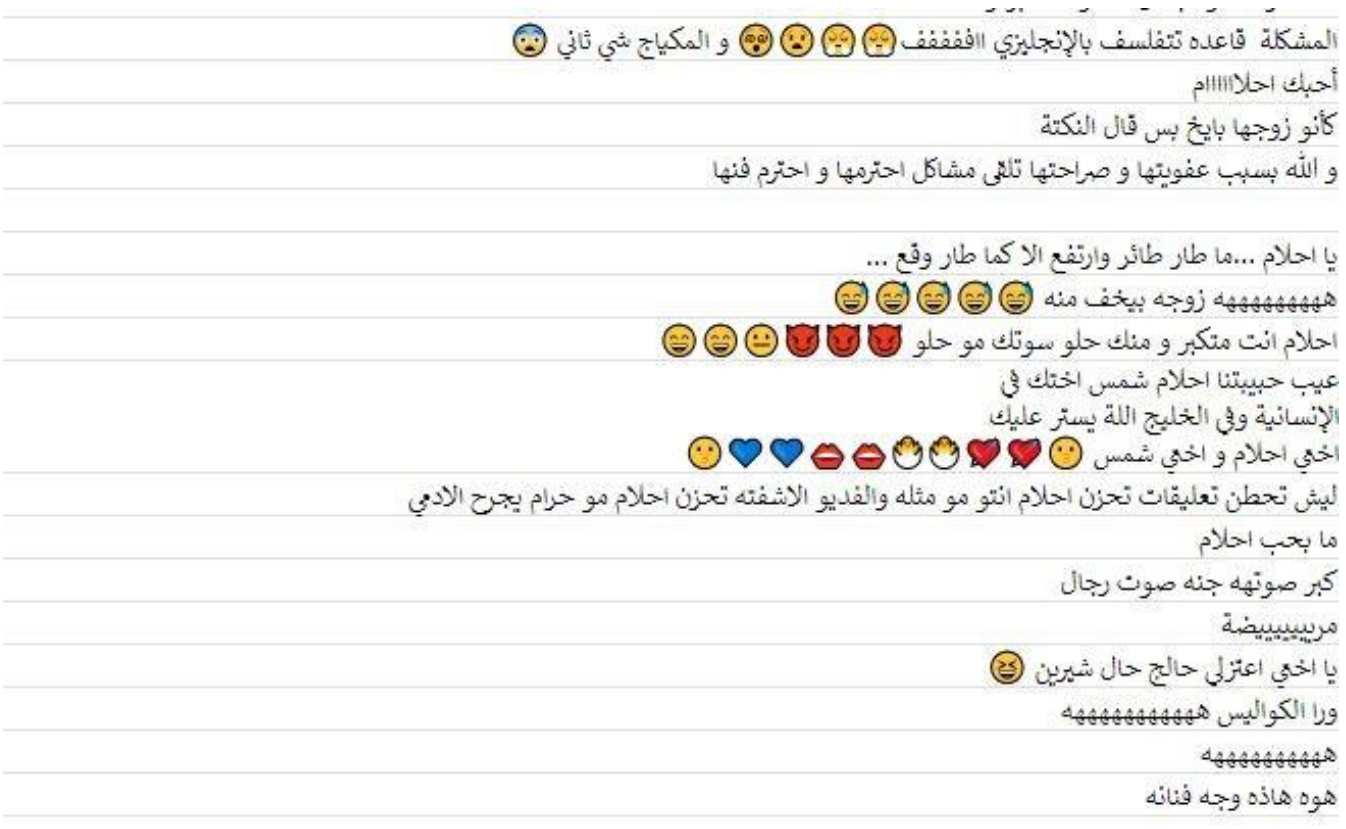

المشكلة قاعده تتفلسف بالإنجليزي اففففف و المكياج شي ثاني
أحبك احلاااام
كأنو زوجها بايخ بس قال النكتة
و الله بسبب عفويتها و صراحتها تلقى مشاكل احترمها و احترم فنها

يا احلام ...ما طار طائر وارتفع الا كما طار وقع ...
هههههههههه زوجه بيخف منه
احلام انت متكبر و منك حلو سوتك مو حلو
عيب حبيبتنا احلام شمس اختك في
الإنسانية وفي الخليج الله يستر عليك
اخي احلام و اخي شمس
ليش تحطن تعليقات تحزن احلام انتو مو مثله والفديو الاشفته تحزن احلام مو حرام يجرح الادمي
ما يحب احلام
كبر صوتهه جنه صوت رجال
مريييييضة
يا اخي اعتزلي حالج حال شيرين
ورا الكواليس هههههههههه
هههههههههه
هوه هاذه وجه فنانه

Figure 3: Snapshot of YouTube dataset.

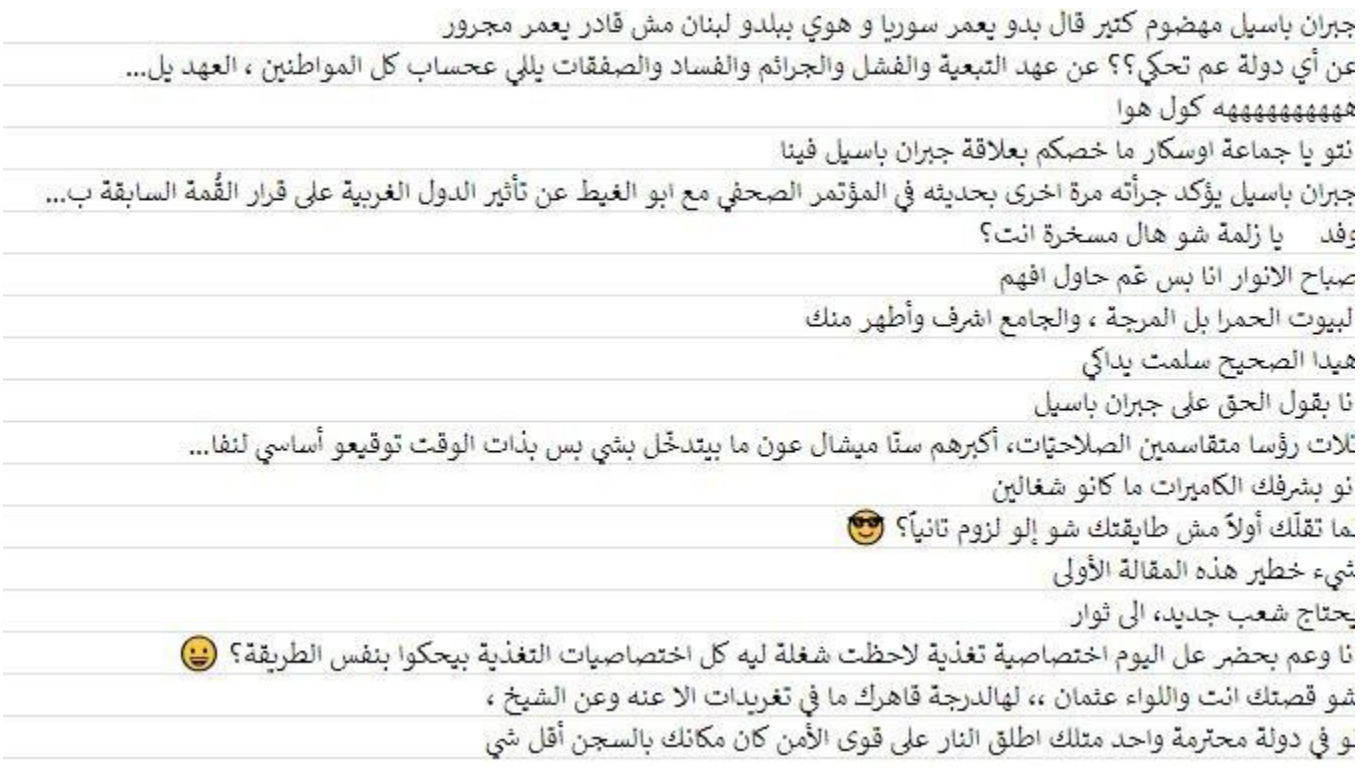

جبران باسيل مهضوم كتير قال بدو يعمر سوريا و هوي ببلدو لبنان مش قادر يعمر مجرور
عن أي دولة عم تحكي؟؟ عن عهد التبعية والفشل والجرائم والفساد والصفقات يللي عحساب كل المواطنين ، العهد يل...
هههههههههه كول هوا
نتو يا جماعة اوسكار ما خصكم بعلاقة جبران باسيل فينا
جبران باسيل يؤكد جرأته مرة اخرى بحديثه في المؤتمر الصحفي مع ابو الغيط عن تأثير الدول الغربية على قرار القُمة السابقة ب...
وفد يا زلمة شو هال مسخرة انت؟
صباح الانوار انا بس عم حاول افهم
لبيوت الحمرا بل المرجة ، والجامع اشرف وأطهر منك
هيدا الصحيح سلمت يداكي
نا بقول الحق على جبران باسيل
ثلات رؤسا متقاسمين الصلاحيات، أكبرهم سنا ميشال عون ما بيتدخل بشي بس بذات الوقت توقيعو أساسي لنفا...
نو بشرفك الكاميرات ما كانو شغالين
ما تقلّك أولاً مش طايقتك شو إلو لزوم تانياً؟
شيء خطير هذه المقالة الأولى
يحتاج شعب جديد، الى ثوار
نا وعم يحضر عل اليوم اختصاصية تغذية لاحظت شغلة ليه كل اختصاصيات التغذية بيحكوا بنفس الطريقة؟
شو قصتك انت واللواء عثمان ،، لهالدرجة قاهرك ما في تغريدات الا عنه وعن الشيخ ،
و في دولة محترمة واحد متلك اطلق النار على قوى الأمن كان مكانك بالسجن أقل شي

Figure 4: Snapshot of L-HSAB dataset

provides a broader perspective on how sarcasm might be intertwined with or differentiated from other forms of communication, particularly those that are anti-social or negative.

3. L-HSAB (Levantine Hate Speech and Abusive Language) Dataset: The L-HSAB dataset is instrumental in understanding the landscape of hate speech and abusive language in Arabic, particularly in the Levantine dialect. Figure 4 shows a snapshot of the L-HSAB dataset. Its inclusion aids in refining the detection algorithms to discern sarcasm from potentially similar but contextually different expressions like hate speech or abusive remarks.

### 3.3 Preprocessing

Preprocessing plays a crucial role in the machine learning workflow for classification tasks. This phase requires the modification and improvement of input data to make it suitable and beneficial for use with transfer learning models. The study implemented the following steps:

1. Data cleaning is a pivotal step in data analysis, emphasizing the importance of removing noise and irrelevant information to enhance data quality. Data cleaning is a pivotal step in data analysis, emphasizing the importance of removing noise and irrelevant information to enhance data quality [Nguyen et al., 2023]. This step was applied to the three datasets by removing unnecessary columns, handling missing values, and text data normalization

2. Text data preprocessing is a fundamental step in NLP, enabling the transformation of raw text into a clean, organized format suitable for analysis. This process involves: (I) using regular expressions to identify and remove unwanted characters, spaces, or patterns within text data, (ii) tokenization is the process of breaking down the text into smaller units, such as words or phrases, (iii) removal of stop words, and (iv) performing Stemming which is reducing words to their root form.

3. Splitting the dataset into training, validation, and test sets in two distinct versions: one version includes emojis in the text data, and the other version with emojis excluded.

### 3.4 Performance Metrics

The evaluation of the models is conducted using a set of standard performance metrics in transfer learning:

- Accuracy: this metric measures the proportion of correctly identified instances (both sarcastic and non-sarcastic) among the total instances. It provides an overall effectiveness of the model.

$$Accuracy = \frac{Numberof\ \ CorrectPredictions}{TotalNumberof\ \ Predictions} \tag{1}$$

- Precision and Recall: these metrics evaluate the accuracy of the model in identifying sarcasm. Precision measures the proportion of correctly identified sarcastic instances among all instances identified as sarcastic, while recall measures the proportion of correctly identified sarcastic instances among all actual sarcastic instances.

$$Precision = \frac{TruePositives}{TruePositives + FalsePositives} \tag{2}$$

$$Recall = \frac{TruePositives}{TruePositives + FalseNegatives} \tag{3}$$

- F1 Score: it is the harmonic mean of precision and recall, providing a single metric that balances both. It is particularly useful when the class distribution is imbalanced.

$$F1 = 2 \times \frac{Precision\ \times\ Recall}{Precision\ +\ Recall} \tag{4}$$

## 4 Results and Analaysis

In this work, the investigation focuses primarily on the influence of emojis in classifying offensive language within Arabic. Furthermore, it presents a comprehensive evaluation and interpretation of the outcomes highlighting which machine learning model can be most effective in accurately identifying offensive speech with or without emojis. The results are critical to knowing the nuances and accuracy of the language processing techniques in a linguistically diverse and complex region like the Arabic world.

The comparative analysis, which is presented in Figures 5-7, assesses the influence of emojis on the performance of Arabic language models in classifying offensive content within the SemEval, YouTube, L-HSAB datasets, respectively. This evaluation is segmented into four distinct metrics: accuracy, recall, precision, and F1-score. The performance of the three considered, AraBERT_v2 (v2), AraBERT_v2_Twitter (TW), and multi_dialect_bert_base_arabert (MD) models are investigated for all of these datasets.

### 4.1 Accuracy

In Figures 5 (a), 6 (a), and 7 (a), the accuracy metrics are evaluated across multiple models, including AraBERT_v2 (v2), AraBERT_v2_Twitter (TW), and multi_dialect_bert_base_arabert (MD). Across these figures, the accuracy rates for each model are displayed both with and without the inclusion of emojis in the dataset. In all cases, except for L-HSAB with TW model, the accuracy tends to be higher when emojis are excluded. This trend suggests that emojis may introduce ambiguity or noise that degrades the models' ability to accurately classify text.

### 4.2 Recall

The recall results across the three models: V2, TW, and MD, are examined through Figures 5 (b), 6 (b), and 7 (b), respectively. For all the datasets, it observed that excluding emojis leads to an increase in recall for TW and MD, while in the V2 models with the youtube and SemEval dataset, recall shows a slight improvement when emojis are included. This suggests that emojis may enhance detection for some models but could potentially disrupt others, resulting in missed detection. The V2 model, however, exhibits negligible differences in both cases.

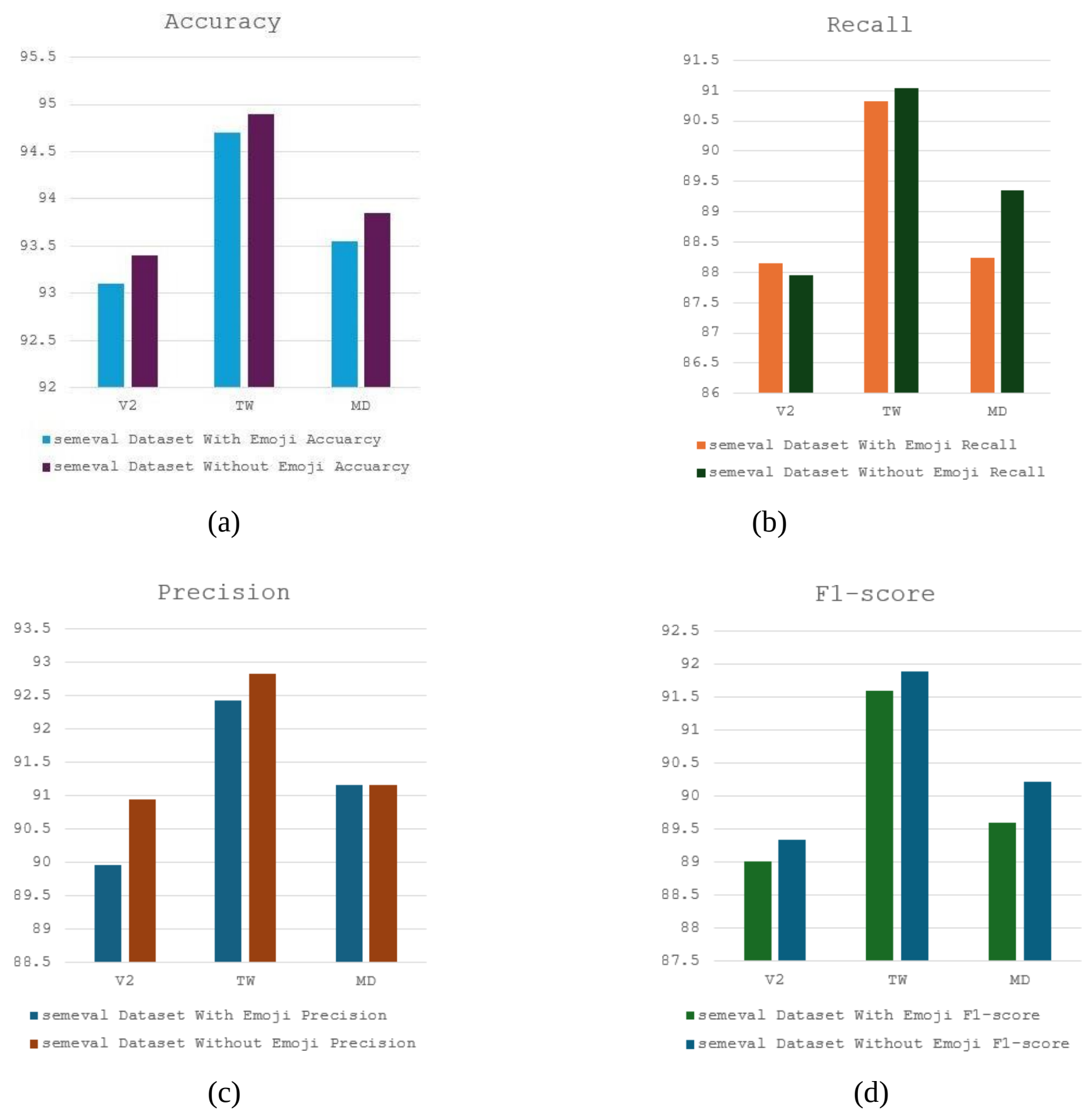

Figure 5: Comparison results of classification with and without emojis on the SemEval dataset.

## 4.3 Precision

Analyzing precision across different models in Figures 5 (c), 6 (c), and 7 (c) yields the following observations. In Figure 5 (c), tThe precision is higher when emojis are excluded for all three models. Particularly, there is a significant increase in precision observed for the V2 and TW models when emojis are excluded, suggesting that emojis may have a notable impact on lowering precision, especially for these models. For Figure 6(c), the precision outcomes show improvements for all models when emojis are removed. The V2 and MD models demonstrate significant enhancement in precision without emojis, indicating a reduction in false positives. The TW model shows no significant difference in both cases. In Figure 7 (c), the MD model shows improvement in precision without emojis, indicating potential complications that emojis introduce in precision tasks across different models. However, the V2 and TW models show a minor increase in precision with emojis, suggesting a better ability to identify positive instances when emojis are present.

## 4.4 F1-score

In Figures 5 and 6(d), a consistent trend is observed where excluding emojis benefits the F1-score across all models. This indicates that emojis do not significantly contribute to the classification process and could potentially even restrict it. The same thing almost exists in Figure 7 (d). For V2, the F1-score slightly drops when emojis are excluded. The difference in the MD case, is almost negligible. These findings implies that emoji exclusion mitigates the trade-off between precision and recall for this particular model.

From these results, the proposed framework can conclude that excluding emojis consistently improves accuracy, recall, precision, and F1-score across various models, indicating its beneficial impact on sarcasm speech classification. Emojis introduce ambiguity or noise that degrades classification accuracy, while their exclusion leads to increased recall, particularly for TW and MD models. Precision notably increases, especially for V2 and MD models, when emojis are removed, reducing false positives. Similarly, F1-score improves across all models when emojis are excluded, except for a slight drop in V2's case, implying a better balance between precision and recall. Therefore, these results support our main hypothesis (RQ1, RQ2 and RQ3) of the positive, or at least the non-negative, impact of removing the emojis in the preprocessing stage on improving sarcasm speech classification performance.

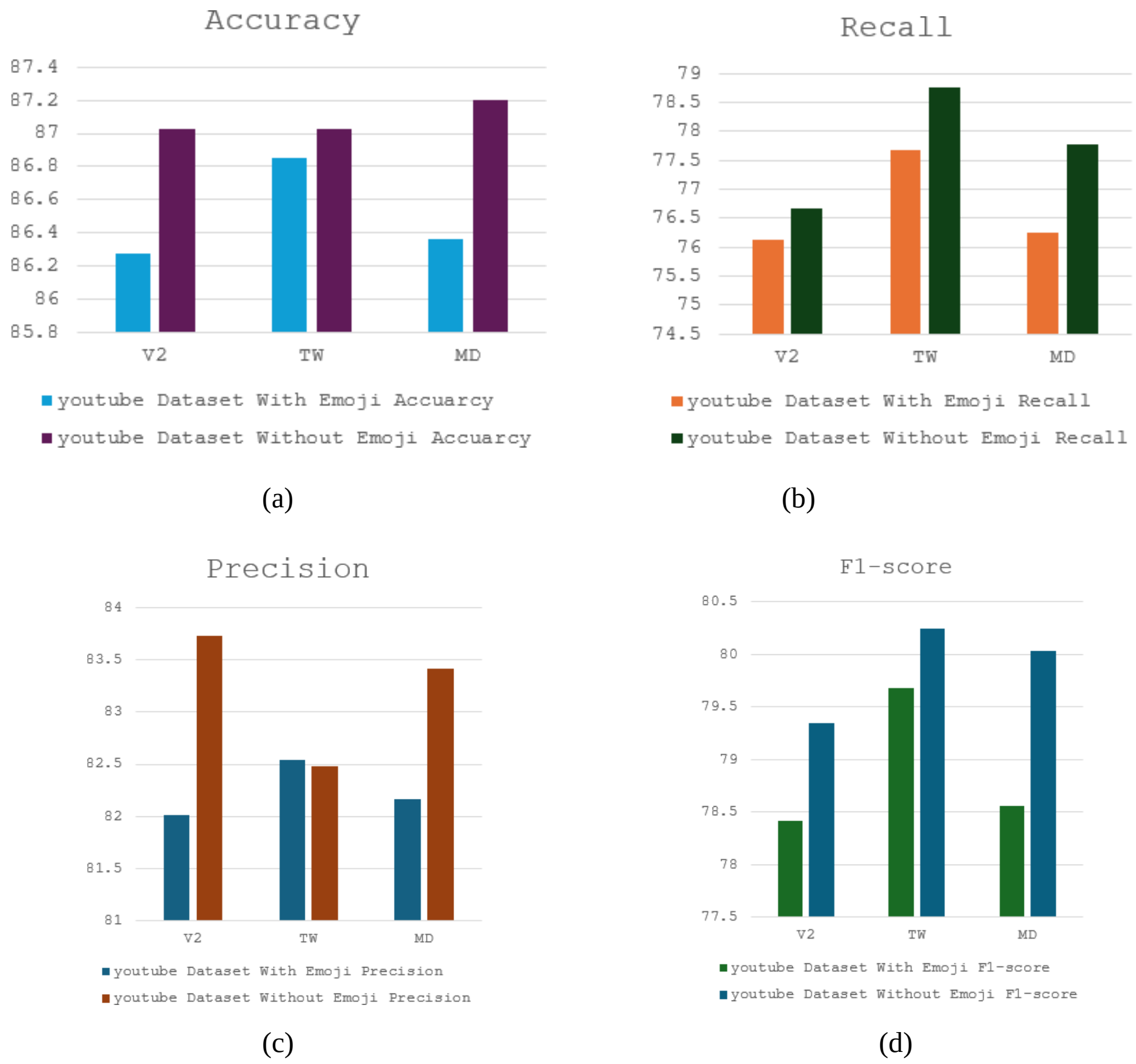

Figure 6: Comparison results of classification with and without emojis on the YouTube dataset.

## 5 Conclusions and Future Work

In this paper, we investigated the impact of excluding emojis during the preprocessing stage in Arabic sarcasm detection models on the performance. Detecting sarcasm in Arabic social media speech presents a multifaceted challenge due to linguistic diversity and the intricacies of sarcastic expressions. By evaluating the accuracy, recall, precision, and F1-score metrics across various models and datasets, we demonstrated that emoji exclusion enhances sarcasm detection accuracy. This research provides a more precise interpretation of language by eliminating potential confusion introduced by non-textual elements, ultimately contributing to the advancement of language processing techniques in linguistically diverse regions. Moreover, our findings provide insights into social media platforms and natural language processing research. By highlighting the positive impact of emoji exclusion on sarcasm detection model performance, new benchmarks are established in Arabic natural language processing.

Possible future research directions might be directed toward developing advanced machine learning tools capable of interpreting emojis in conjunction with text. This will mitigate the negative confusion effect that is caused be misuse or overuse of emojis. Moreover, this research provides foundation for enhancing real-time monitoring tools on social media platforms. Such tools could facilitate sentiment analysis, content moderation, and the tracking of public opinion trends.

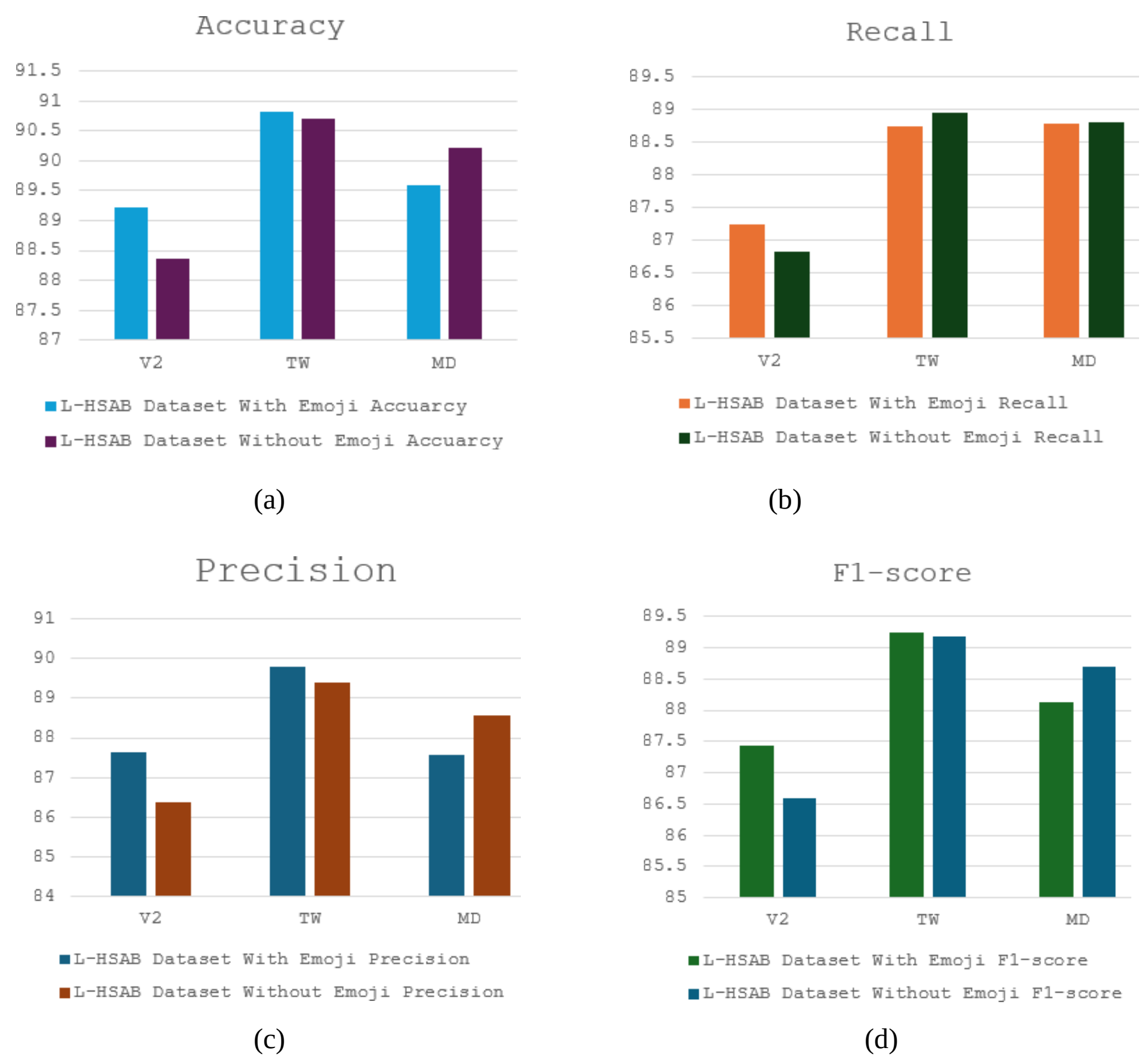

Figure 7: Comparison results of classification with and without emojis on the L-HSAB dataset.